# Attacking and Defending Covert Channels and Behavioral Models

Valentino Crespi, George Cybenko, Annarita Giani

## Abstract

In this paper we present methods for attacking and defending $k$-gram statistical analysis techniques that are used, for example, in network traffic analysis and covert channel detection. The main new result is our demonstration of how to use a behavior's or process' $k$-order statistics to build a stochastic process that has those same $k$-order stationary statistics but possesses different, deliberately designed, $(k+1)$-order statistics if desired. Such a model realizes a "complexification" of the process or behavior which a defender can use to monitor whether an attacker is shaping the behavior. By deliberately introducing designed $(k+1)$-order behaviors, the defender can check to see if those behaviors are present in the data. We also develop constructs for source codes that respect the $k$-order statistics of a process while encoding covert information. One fundamental consequence of these results is that certain types of behavior analyses techniques come down to an *arms race* in the sense that the advantage goes to the party that has more computing resources applied to the problem.

Points of view in this document are those of the authors and do not necessarily represent the official position of the sponsoring agencies or the U.S. Government.

V. Crespi is with the Department of Computer Science, California State University at Los Angeles, Los Angeles CA, 90032 USA. email: vcrespi@calstatela.edu. Crespi's work was partially supported by AFOSR Grant FA9550-07-1-0421 and by NSF Grant HRD-0932421.

G. Cybenko is with the Thayer School of Engineering, Dartmouth College, Hanover NH 03755. email: gvc@dartmouth.edu. Cybenko's work was partially supported by Air Force Research Laboratory contracts FA8750-10-1-0045, FA8750-09-1-0174, AFOSR contract FA9550-07-1-0421, U.S. Department of Homeland Security Grant 2006-CS-001-000001 and DARPA Contract HR001-06-1-0033

A. Giani is with the Department of EECS, University of California at Berkeley, Berkeley CA 94720. email: agiani@eecs.berkeley.edu. Giani's's work was partially supported by U.S. Department of Homeland Security Grant 2006-CS-001-000001 and DARPA Contract HR001-06-1-0033 when she was a Ph.D. student at Dartmouth

**Index Terms**

Covert Channels, Exfiltration, Probabilistic Automata, Cognitive Attack, Anomaly Detection.

## I. Introduction

Computer security researchers have been investigating statistical behavioral modeling techniques as a means for determining whether a machine, a network or data packet contents are behaving "normally" or not. These are so-called behavior analysis techniques and implicitly model stochastic processes at some level of fidelity.

Consider for example, the problem of detecting covert channels. Some existing approaches assume that an adversary has installed an exfiltrating agent, or Trojan, which operates by encoding data in a way that introduces detectable regularities in some network traffic statistics. For example, Giani et al. [1] and Cabuk et al. [2] estimate certain first order statistics of packet inter-arrival delays in order to determine whether a time covert channel is being used. Dainotti et al. [3] learn a Hidden Markov Process [4], [5], [6], [7], [8] using both packet inter-arrival delays and packet sizes to detect traffic anomalies. Other techniques are based on various analyses of $n$-gram statistics [9]. In fact, some have called techniques that match $n$-gram statistics "mimicry attacks" and while techniques have been developed for detecting certain simple types of mimicry, techniques for building mimicry attacks as described in the present paper appear to be novel [9].

General discussions of covert channels and their taxonomies, existence and modeling have been published [10], [11], [12], [13], [14]. The design, implementation and experimental evaluation of several specific covert channel attacks in real systems is of specific interest [14]. That work presents threat models, achievable bit rates, noise properties and channel capacities for covert channels.

The existence and successful use of a covert channel is based on the assumption that the covert channel code does not perturb the measured statistical properties of behavior so that, over time, a covert transmission does not introduce discernible patterns which are different than expected, at least with respect to what is measured. In this paper we assume the ability to learn a $k$-gram type model of "normal behavior." This is simply done by counting the occurrences of $k$-grams and then normalizing to produce frequencies or probabilities. It is important to note that researchers often talk about entropy as a channel statistic [10], [15] but entropy is typically calculated from $k$-order statistics so that our methods for preserving $k$-order statistics preserves all lower order statistics and will also preserve the entropy.

We present a technique for encoding messages that respects these $k$-order statistics. Both attacker and defender can use this coding technique. The attacker could exfiltrate coded information while the defender could embed an encoded reference message or carrier to detect manipulations of the channel by an adversary attempting covert communications. That is, for any order $k$, an attacker or a defender can encode covert messages while otherwise respecting the $k$-order statistics of the traffic.

Also, we show how a defender can create a process of order $k + 1$ which has the same $k$-order statistics but specifically designed $(k + 1)$-order statistics that the defender can easily monitor to see if the $(k+1)$-order statistics have been changed. Researchers have recently started to develop systematic taxonomies and examples of attacks against statistical machine learning techniques [16]. In that spirit, the present work develops specific techniques to both attack and defend using certain statistical approaches.

We discuss these methods in the context of behaviors that have a finite set of observable symbols (the alphabet). Interpacket arrival times, packet sizes, header fields, packet contents and so on are examples of such observables if quantized into a finite number of bins. Our approach models the observables as a stationary stochastic process $\mathcal{X}$ [17]. After estimating the $k$-order statistics, we build a Probabilistic Deterministic Finite Automata model (PDFA) [18], [19], [20], [21] that realizes the $k$-order statistics.

Using that PDFA, we show that:

1) an adversary can encode messages covertly while respecting the $k$-order statistics;

2) the defender can encode reference messages or a carrier while respecting the $k$-order statistics and;

3) the defender can build a more complex process which has the same $k$-order statistics but possesses deliberately designed $(k + 1)$-order statistics.

Examples of such covert channels in network traffic include, but are not restricted to:

- *Timing Channels:* The observable symbols are the inter-packet time delays, appropriately quantized;

- *Size Channels:* The observable symbols are quantized sizes of the packets;

- *Header Channels:* The observable symbols are various header fields in TCP/IP packets

which can be manipulated by the transmitting entity without violating protocol semantics. Several such fields are known to exist [22].

It is important to clarify right away what we mean by a $k$-order statistic and a $k$-gram. Suppose we have an alphabet consisting of $\{\alpha, \beta, \gamma\}$ and we observe a sequence comprised of that alphabet, say

$$\alpha\alpha\alpha\beta\alpha\beta\gamma\gamma.$$

The first order statistics are $[1/2 \ 1/4 \ 1/4]$ indicating that 1/2 of the symbols are $\alpha$'s, 1/4 are $\beta$'s and 1/4 are $\gamma$'s. The 1-grams are merely $\{\alpha, \beta, \gamma\}$.

The 2-grams observed in this sequence are $\alpha\alpha$, $\alpha\alpha$, $\alpha\beta$, $\beta\alpha$, $\alpha\beta$, $\beta\gamma$, $\gamma\gamma$ and the 2-order statistics for the 9 possible 2-grams

$$\alpha\alpha, \ \alpha\beta, \ \alpha\gamma, \ \beta\alpha, \ \beta\beta, \ \beta\gamma, \ \gamma\alpha, \ \gamma\beta, \ \gamma\gamma$$

are respectively

$$[2/7 \ 2/7 \ 0 \ 1/7 \ 0 \ 1/7 \ 0 \ 0 \ 1/7].$$

That is, our $k$-grams are obtained by moving a sliding window of width $k$ across the data one symbol at a time. This is not to be confused with moving that window across the data sequence $k$ symbols at a time.

The following discussion shows how a timing covert channel can be constructed based on a beacon and argues that a naive encoding of covert messages based on packet inter-arrival times produces a clearly detectable distortion of the 1-order statistics of those time intervals in network traffic [23], [1].

Figure 1 describes the setup. Machine A sends a regularly timed beacon to machine D. (Such a beacon can be a time server request or a *stay alive* beacon for instance.) The inter-packet delays seen at machine B are not regular due to internal routing delays in the LAN. (These statistics were actually measured from a regularly timed beacon traveling several hops.) An intruder was able to compromise and control machine B which is inside the local network and a relay for the traffic between A and D. (B could be a proxy server, border router or other device for example.) Assume we set up a machine C outside the internal network perimeter to check for timing covert channels. C has seen a certain distribution of inter-packet delays coming from A going to D.

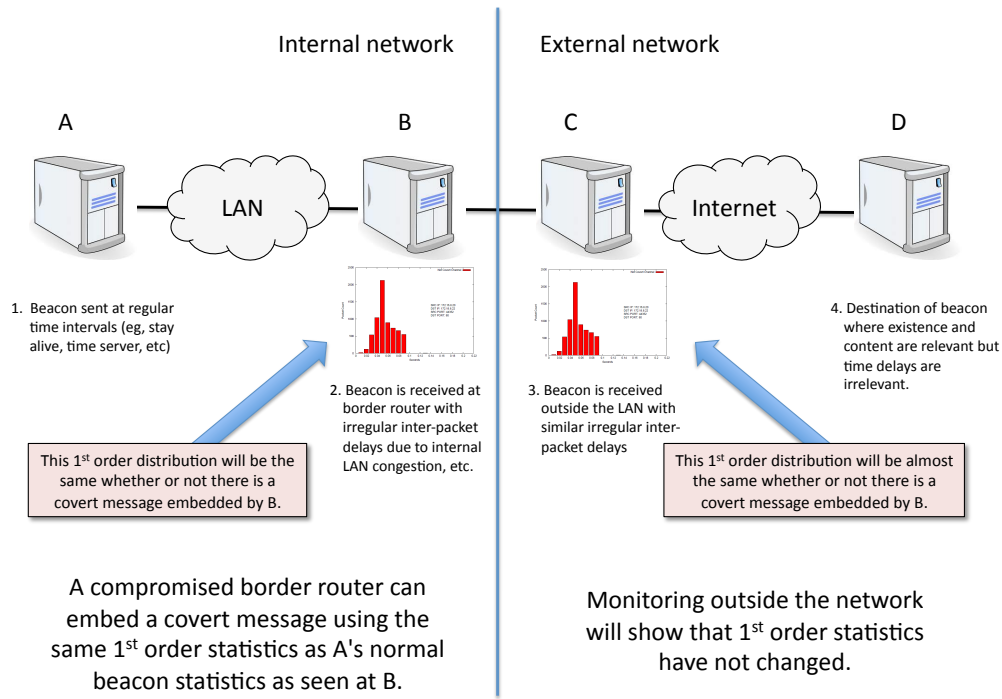

Fig. 1. An intruder controlling machine B inside a local network exfiltrates data coded in inter-packet delays received by machine C en route to machine D. Monitoring outside the intranet LAN will show the same first order inter-packet delays with or without the covert channel as constructed in this paper.

In this paper, we show how machine B can encode covert messages in the inter-packet delays in such a way that the first order statistics as seen by C remain unchanged from the original distribution. Conversely, we can deliberately defend against such channels by encoding messages so that any manipulation of the delays will be detectable on the outside at machine C, because the covert message will not be received at C.

Figure 2 shows the number of packets received with a given delay in two scenarios. The horizontal axis reports the inter-arrival time in seconds, and the vertical axis the number of packets received with those delays. In the left graph of Figure 2 are the observed inter-packet delays resulting from a regularly timed beacon traversing multiple hops in a LAN. On the right hand graph, we depict a naive covert timing channel using two time intervals to encode a message. It is evident from the data that the naive covert communication in the right graph can be easily detected if the 1-order statistics of normal traffic have been measured and are those on the left. However, the 1-order distribution on the left can be generated either by normal traffic, as it was obtained, or by a covert channel, as we will show.

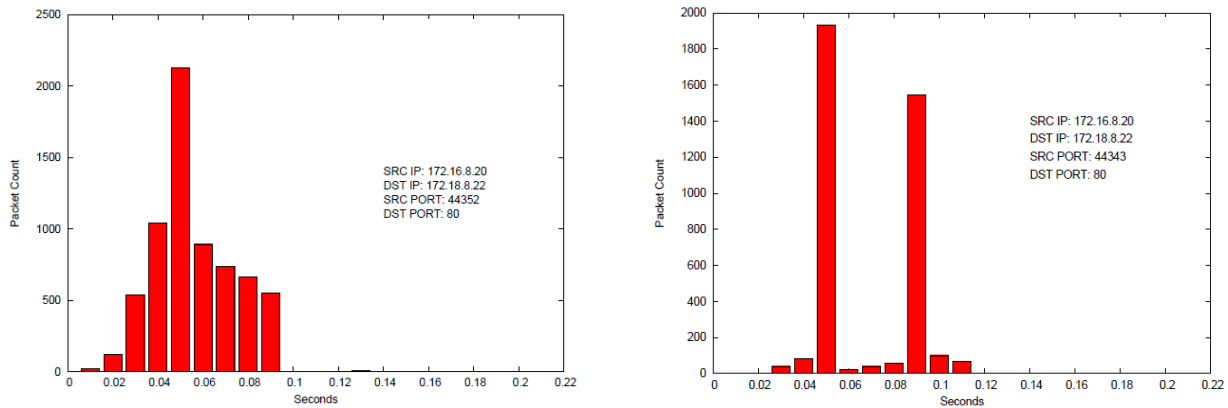

Fig. 2.    First order statistics of inter-packet delays of normal traffic (left) and a poorly designed covert channel using two delays only (right). (Packets were deliberately routed to hop several times between source and destination.) This paper develops techniques for creating covert channels that have the same statistics as the ones depicted on the left, even if higher order statistics are measured.

In this contribution, we develop a more sophisticated approach than the naive approach insofar as we consider also statistics of arbitrarily higher order, i.e. $k > 1$, and our results effectively show that, for any $k$, defenders and attackers both have technical approaches for, respectively, attacking or defending a $k$-order behavior with respect to covert communications. Consequently, the situation is an *arms race* in the sense that whichever side has the ability to learn the highest order statistics wins.

### A.  Outline of the Paper

In Section II we present an illustrative example. In Section III we describe our method and show how to manipulate a behavior's statistics with Probabilistic Automata. In Section IV we provide a numerical example. In Section V we show how to use Probabilistic Automata to build a channel code that respects the statistics of traffic up to some predecided order. Finally, Section VI contains some conclusions and future work references.

## II. A Simple Illustrative Example

To illustrate these concepts, consider a simple binary observable with values, 0 and 1. It is assumed that these observables are irrelevant to the normal operation of the underlying system and its semantics. For example, the observables could be quantized inter-arrival times or unused packet header fields.

Assume that the 1-order statistics of these observables are $r_0 > 0$ and $r_1 > 0$ with $r_0 + r_1 = 1$. This means that the relative frequency of 0's and 1's as observed in the behavior are $r_0$ and $r_1$ respectively. Now suppose an attacker has estimated these probabilities and seeks to exfiltrate messages while respecting these probabilities. This is possible and, later in this paper, we review standard source coding ideas that allow the attacker to create such codes efficiently.

In fact, if the messages to be sent are binary and Bernoulli with $p = 0.5$ (such as for encrypted and/or compressed messages), then there are codes that use $1/H(r_0) = 1/H(r_1)$ bits in the covert channel per original message symbol where $H(x) = -(x \log_2 x + (1 - x) \log_2(1 - x))$ is the entropy function. We show how to construct such codes to respect $k$-order statistics as well.

By the same token, the defender can encode a reference signal, also respecting the first order statistics as above, which can be decoded and verified at the receiving end.

Note that no specific second order statistics $r_{00}, r_{01}, r_{10}$ and $r_{11}$ have been modeled so far, but if the process is modeled by a Bernoulli process with $p = r_1$ then the second order statistics would be $r_i \cdot r_j = r_{ij} = Prob(ij) = Prob(ji)$ by independence.

However, the defender can construct a second order process with second order statistics $r_{00}, r_{01}, r_{10}$ and $r_{11}$ for which $r_{ij} \neq r_i \cdot r_j$ while satisfying the required first order statistics, namely $r_0$ and $r_1$. If the attacker exploits the channel through a purely first order process, the constructed second order statistics $r_{ij}$ will likely not be observed by the defender who could then conclude that the traffic is being shaped by an adversary.

To illustrate this 2-order construction, consider an automaton with two states, $Q = \{0, 1\}$, corresponding to the two 1-grams of observables. Let $X$ be the matrix of the transition probabilities

$$X = \begin{bmatrix} p_{00} & p_{01} \\ p_{10} & p_{11} \end{bmatrix}.$$

We seek PDFAs that have the stationary distribution $\pi = [r_0 \quad 1 - r_0] = [1 - r_1 \quad r_1] = [1 - r \quad r]$.

Specifically, we seek $X$ that satisfies

$$[1 - r \quad r] \cdot X = \pi X = \pi = [1 - r \quad r]$$

with $X$ being a stochastic matrix (non-negative with row sums equal to 1). The class of PDFAs that are 1-order equivalent to the given process is therefore determined by a set of linear equality and inequality constraints as follows:

$$
\begin{aligned}
(1 - r) \cdot p_{00} + r \cdot p_{01} &= 1 - r \\
(1 - r) \cdot p_{10} + r \cdot p_{11} &= r \\
p_{00} + p_{01} &= 1 \\
p_{10} + p_{11} &= 1 \\
p_{ij} &\geq 0.
\end{aligned}
$$

The four equations are linearly dependent and we can reduce them to the three equations and constraints

$$
\begin{aligned}
r \cdot p_{11} - (1 - r) \cdot p_{00} &= 2 \cdot r - 1 \\
p_{00} + p_{01} &= 1 \\
p_{10} + p_{11} &= 1 \\
0 \leq p_{00} , \ p_{11} &\leq 1.
\end{aligned}
$$

There are an infinite number of solutions according to

$$p_{11} = \frac{1 - r}{r} p_{00} + \frac{2r - 1}{r}, \quad 0 \leq p_{00} , p_{11} \leq 1. \tag{1}$$

For example, if $r = 0.3, \ 1 - r = 0.7$ then the constraints become:

$$
\begin{aligned}
p_{11} &= \frac{0.7 p_{00} - 0.4}{0.3} \\
0 \leq p_{00} , \ p_{11} &\leq 1
\end{aligned}
$$

so letting $p_{00} = 0.8$ we get $p_{11} = \frac{0.16}{0.3} = 0.5\bar{3}$ and therefore $p_{01} = 0.2$ and $p_{10} = 0.4\bar{6}$. This yields 2-order statistics of:

$$
\begin{aligned}
r_{00} &= p_{00}\pi_1 = 0.8 \cdot 0.7 = 0.56 \\
r_{01} &= p_{01}\pi_1 = 0.2 \cdot 0.7 = 0.14 \\
r_{10} &= p_{10}\pi_2 = 0.4\bar{6} \cdot 0.3 = 0.14 \\
r_{11} &= p_{11}\pi_2 = 0.5\bar{3} \cdot 0.3 = 0.16.
\end{aligned}
$$

Notice that $r_{01} = r_{10} = 0.14$, $r_{01} + r_{00} = r_0 = \pi_1 = 0.3$ and $r_{01} + r_{11} = r_1 = \pi_2 = 0.7$ as required.

Another, equivalent way to derive these relations is to note that there are two trivial solutions for $X$, namely $X_1 = I_2$ (the 2 by 2 identity matrix) and $X_2 = \mathbf{1} \cdot \pi$ where $\mathbf{1} = [1\ 1]^T$ is the column vector whose entries are all 1's. These two solutions are always different. Moreover, we can see that any convex combination $\rho X_1 + (1 - \rho)X_2$ for $0 \le \rho \le 1$ is also a solution to all the constraints and in fact yields the same class of solutions as above.

The point of this example is that we can shape the second order statistics of the observables without changing the first order statistics. In particular, multiple choices for $p_{00}$ (and so for $r_{00}$) are possible, all of which lead to the same 1-order statistics. A defender can shape the second order statistics so that if an attacker only obeys the first order statistics, the defender can detect that the expected second order statistics are wrong.

Note that the second order process in this example satisfies an additional constraint - namely, the marginal distributions must agree with the first order process, namely $r_{01} + r_{11} = r_1$ and so on. Moreover, $r_{01} = r_{10}$ must be true as well (this is a symmetry which arises from considering the $0$ to $1$ and $1$ to $0$ transitions in the observed sequence which must be equal). For higher order processes, the construction involves identifying and dealing with additional constraints and finding realizations which satisfy them. These generalizations to higher orders are one of the main contributions of this paper.

To apply this construction to the empirical data shown in Figure 2, normalize the counts into frequencies or probabilities by dividing by the total packet count. This yields a vector of probabilities:

$$
R = [\ 0.0029\ 0.0144\ 0.0734\ 0.1453\ 0.3094\ 0.1295\ 0.1151\ 0.1079\ 0.1007\ 0\ 0\ 0\ 0.0014\ ] \quad (2)
$$

where the coordinates 1 through 13 correspond to delays of 0.01 through 0.13 in increments of 0.01.

We seek to construct a Markov Chain whose states correspond to observable inter-packet delays and whose transition probabilities, $P$, describe the probability that one delay follows another. As explained above, $P$ must satisfy two matrix equations (capturing the facts that $R$ is a stationary vector for $P$ and that $P$ is row stochastic)

$$R * P = R \text{ and } P * \mathbf{1} = \mathbf{1}$$

where $\mathbf{1}$ is the column vector of all ones. Moreover, the entries of $P$ are all non-negative.

In this simple case, there are two solutions which are simple to identify, namely

$$P_B = \mathbf{1} * R \text{ and } P_D = I \tag{3}$$

where $I$ is the 13 by 13 identity matrix. The reader can easily check that both these matrices satisfy the two required matrix equations. This construct is simple for 1-grams but becomes more complex for general $k$-grams as shown below.

Moreover, for any $0 \le \alpha \le 1$, $P_\alpha = \alpha P_B + (1 - \alpha) P_D$ is also a solution. Whereas $P_B$ defines a Bernoulli process and $P_D$ describes a completely disconnected Markov Chain with an infinite number of fixed distributions, $P_\alpha$ defines a Markov Chain that is irreducible, aperiodic and not a Bernoulli process for any $0 < \alpha < 1$ . Therefore, $P_\alpha$ can be used by a defender to create specific second order statistics which an attacker would have to first model and then respect.

## III. Constructing the Automata

In this section, we show how to construct automata that can reproduce observed statistics computed from data.

Let $\Sigma = \{a, b, c, ...\}$ be the finite observable alphabet and $\sigma = |\Sigma| < \infty$ be the number of observables. We are assuming that we have sequences of observables from which we compute the relative frequencies of $k$-grams ($k \ge 1$):

$$0 \le R(x) \le 1, \quad \sum_{x \in \Sigma^k} R(x) = 1.$$

Here $\Sigma^k$ is the set of $k$-grams; that is, the set of all possible sequences of length $k$ drawn from the alphabet $\Sigma$.

Roughly speaking, if $s_0 s_1 ... s_{n-1} = S_{0:n-1}$ is an observed data sequence of length $n > k$, $R(x)$ is approximated by the number of occurrences of the substring $x$ in $S_{0:n-1}$ divided by the total number of substrings of length $k$ in $S_{0:n-1}$, namely $n - k + 1$. The set of $R(x)$'s is precisely what we mean by the $k$-order statistics of the observations.

These statistics must satisfy certain regularity conditions required by the proposed construction so some care must be taken in their computation. Specifically, the identity

$$\sum_{a \in \Sigma} R(ay) = \sum_{b \in \Sigma} R(yb) = R(y)$$

should hold for every $y \in \Sigma^{k-1}$. This can be accomplished by appending $S_{0:n-1}$ with $s_0 s_1 ... s_{k-2}$ as a suffix, creating a periodic string effectively, and counting occurrences in the periodic string.

Moreover, this can be repeated for every $1 < j < k$ by using a circular buffer appending $s_0 s_1 ... s_{j-2}$. All marginal distributions

$$\sum_{w \in \Sigma^{k-j}} R(wy) = \sum_{w \in \Sigma^{k-j}} R(yw) = R(y)$$

will hold for all $y \in \Sigma^j$ then. (Details are left to the reader.)

We will now construct a special type of Markov Chain in which $\Sigma^k$ are the states and the semantics of the $k$-grams are preserved so that if $x = ay \in \Sigma^k$ is an observed $k$-gram, then $P(ay, yb)$ is the probability of transitioning to state $yb$ where both $a$, $b \in \Sigma$. Such transitions are the only ones possible in the Markov Chain $k$-gram model. Such models are called $k^{\text{th}}$-order Markov Models, $k$ Markov Chains or $k$-gram models by different authors [19], [24].

Let $\pi$ be the vector of measured $k$-gram statistics, $R(x)$, and let $P$ be the desired Markov Chain transition probabilities:

$$P = (P(x, x'))$$

where the entries of both $\pi$ and $P$ are indexed by $x$, $x' \in \Sigma^k$.

The stationary probabilities of the desired Markov Chain are precisely $\pi$ when the equation $\pi P = \pi$ is satisfied. This matrix equation consists of $\sigma^k$ equations and the stochasticity requirement on $P$ is another $\sigma^k$ equations resulting in the following $2\sigma^k$ equations overall:

$$\sum_{x \in \Sigma^k} P(x, x') R(x) \;=\; R(x'), \;\; \forall x' \in \Sigma^k \;, \;\; \text{(stationary probability conditions)} \qquad (4)$$

$$\sum_{x' \in \Sigma^k} P(x, x') \;=\; 1, \;\; \forall x \in \Sigma^k \;\; \text{(probability requirements)} \qquad (5)$$

where $P(x, x') \geq 0$ as well.

Because of the relationship between $k$-grams and the Markov Chain that we are seeking to construct, we can only have $P(x, x') \neq 0$ when $x = ay$ and $x' = yb$ for some $a, b \in \Sigma$ and $y \in \Sigma^{k-1}$. That is, $y$ is the suffix of the state $x = ay$ and we can only transition to states $x' = yb$ which have $y$ as a prefix and some suffix $b \in \Sigma$. Accordingly, for every $y \in \Sigma^{k-1}$, we have the $2\sigma$ equations

$$\sum_{a \in \Sigma} P(ay, yb) R(ay) = R(yb), \quad \forall b \in \Sigma \ , \tag{6}$$

$$\sum_{b \in \Sigma} P(ay, yb) = 1, \quad \forall a \in \Sigma \tag{7}$$

$$P(ay, yb) \geq 0 \tag{8}$$

which are completely decoupled from the equations corresponding to $(k-1)$-grams other than $y$. Accordingly, we can solve each system independently.

Noting that the $k$-grams statistics, $R(x)$, satisfy the marginalization relations

$$\sum_{a \in \Sigma} R(ay) = \sum_{b \in \Sigma} R(yb) = R(y), \quad \forall y \in \Sigma^{k-1},$$

summing over $b$ in the equations (6), we get

$$\sum_{b \in \Sigma} \sum_{a \in \Sigma} P(ay, yb) R(ay) = \sum_{a \in \Sigma} \sum_{b \in \Sigma} P(ay, yb) R(ay) = \sum_{a \in \Sigma} R(ay) = \sum_{b \in \Sigma} R(yb) = R(y)$$

which is an identity not involving the unknown $P(ay, yb)$.

Accordingly, there are no more than $2\sigma - 1$ linearly independent equations in (6). In fact, if we define $pre(y)$ to be the number of nonzero $R(ay)$ and $post(y)$ be the number of nonzero $R(yb)$, there are in fact no more than $pre(y) \cdot post(y)$ unknown probabilities, $P(ay, yb)$, and no more than $pre(y) + post(y) - 1$ independent equations altogether.

### A. The Standard Solution

One solution to the equations, which we call the Standard Solution, is $\bar{P}(ay, yb) = R(yb)/R(y)$ because then

$$\sum_{a \in \Sigma} \bar{P}(ay, yb) R(ay) = \sum_{a \in \Sigma} R(ay) R(yb)/R(y) = R(yb)/R(y) \sum_{a \in \Sigma} R(ay) = R(yb)$$

and

$$\sum_{b \in \Sigma} \bar{P}(ay, yb) = \sum_{b \in \Sigma} R(yb)/R(y) = R(y)/R(y) = 1.$$

This specific solution has $pre(y) \cdot post(y)$ nonzero probabilities, $P(ay, yb)$, for the substring $y \in \Sigma^{k-1}$ by construction.

By construction, this Markov Chain is irreducible because we have constructed the transition probabilities from a circular buffer so that there is a nonzero probability of going from any state with nonzero probability, namely $R(x)$, to any other state with nonzero probability. If additionally the constructed Standard Solution Markov Chain is aperiodic, its unique stationary distribution is precisely $R(x)$ and its entropy rate is

$$H(\bar{P}) = H_{\overline{P}}(X_{k+1}|X_1^k) = -\sum_{a \in \Sigma} \sum_{y \in \Sigma^{k-1}} \sum_{b \in \Sigma} R(ay)\bar{P}(ay, yb) \log(\bar{P}(ay, yb)). \tag{9}$$

### B. Extended Solutions

If $pre(y)$ and $post(y)$ are both strictly greater than 1, then $pre(y) \cdot post(y) > pre(y) + post(y) - 1$. From the theory of linear programming, there are feasible solutions to the linear program defined by (6), (7) and (8) which have no more than $pre(y) + post(y) - 1$ nonzero coordinates, namely the Basic Feasible Solutions [25].

Let such a Basic Feasible Solution be $\hat{P}(ay, yb)$. As derived above, there are solutions with exactly $pre(y) \cdot post(y)$ nonzero coordinates, namely the Standard Solutions, $\bar{P}(ay, yb)$. Note that strict convex combinations of $\hat{P}$ with $\bar{P}$, $P_u = u\hat{P} + (1 - u)\bar{P}$ with $0 < u < 1$, define a continuum of solutions to (6), (7) and (8), with each solution corresponding to an irreducible Markov Chain. This is the case because every state is reachable from every other state with nonzero probability due to the construction of the Standard Solution.

Moreover, when $pre(y)$ and $post(y)$ are both strictly greater than 1, $\hat{P}$ and $\bar{P}$ are different. As an aside, we have observed that Basic Feasible Solutions typically result in reducible chains because those solutions involve a minimal number of nonzero transition probabilities.

## IV. NUMERICAL EXAMPLES

In this section we demonstrate the constructions described above.

1) We consider data generated by the automata depicted in Figure 3 which is a Hidden Markov Model (HMM), $M = \{A(0), A(1)\}$, defined by the two transition matrices

$$A(0) = \begin{bmatrix} 0.5 & 0.5 \\ 0 & 0.5 \end{bmatrix} , \quad A(1) = \begin{bmatrix} 0 & 0 \\ 0.5 & 0 \end{bmatrix} .$$

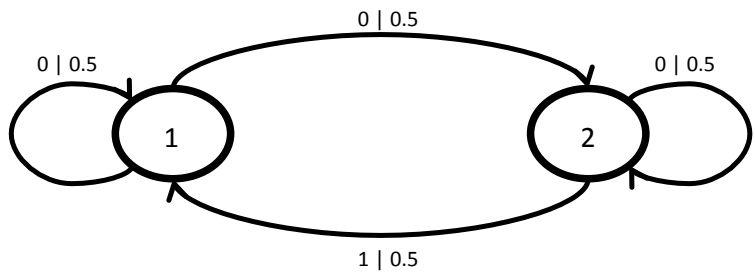

Fig. 3. A two state Hidden Markov Model used to generate data for the example. Transitions between states are labeled with the emitted symbols and the probabilities that the transitions occur so that 0 | 0.5 means the transition occurs with probability 0.5 and emits the symbol "0."

The stochastic process of observables is not Markovian of any order as can be seen by the fact that

$$P(y_t = 0 | y_{t-k}...y_{t-1} = 0^k) \neq P(y_t = 0 | y_{t-k-1}y_{t-k}...y_{t-1} = 10^k)$$

for any $k$. Moreover, it can be shown that this process is not equivalent to any Probabilistic Deterministic Finite State Automaton.

We generated a sequence of 1000 observations by performing a simulation of this HMM, starting in state 1.

2) We set $k = 2$ and computed the statistics $R(xy)$ and $R(xyz)$ by scanning the data sequence from left to right and computing sample averages as appropriate:

$$\begin{bmatrix} R(00) & = & 0.513 \\ R(01) & = & 0.244 \\ R(10) & = & 0.244 \\ R(11) & = & 0.000 \end{bmatrix},$$

and

$$\begin{bmatrix} R(000) & = & 0.338 \\ R(001) & = & 0.174 \\ R(010) & = & 0.244 \\ R(011) & = & 0.000 \\ R(100) & = & 0.174 \\ R(101) & = & 0.070 \\ R(110) & = & 0.000 \\ R(111) & = & 0.000 \end{bmatrix} \ .$$

Observe that $R(01) = R(10)$ which is a necessary regularity that follows from the marginalization property:

$$\sum_a R(ay) = \sum_b R(yb) = R(y) \ .$$

In order to be sure that the estimates verify those consistency conditions we have treated the data stream as a circular buffer as described previously.

3) We built the Standard Solution, $\overline{P}$, where $\overline{P}(ay, yb) = R(yb)/R(y)$, and then we computed a different numerical solution, $\hat{P}$, of the linear program (6), (7) and (8).[1] The two solutions are summarized below:

$$\begin{bmatrix} \overline{P}(00,00) & = & 0.678 \\ \overline{P}(00,01) & = & 0.322 \\ \overline{P}(10,00) & = & 0.678 \\ \overline{P}(10,01) & = & 0.322 \\ \overline{P}(01,10) & = & 1.000 \\ \overline{P}(01,11) & = & 0.000 \\ \overline{P}(11,10) & = & 1.000 \\ \overline{P}(11,11) & = & 0.000 \end{bmatrix}, \begin{bmatrix} \hat{P}(00,00) & = & 1.000 \\ \hat{P}(00,01) & = & 0.000 \\ \hat{P}(10,00) & = & 0.000 \\ \hat{P}(10,01) & = & 1.000 \\ \hat{P}(01,10) & = & 1.000 \\ \hat{P}(01,11) & = & 0.000 \\ \hat{P}(11,10) & = & 0.000 \\ \hat{P}(11,11) & = & 1.000 \end{bmatrix} \ .$$

Note that the Basic Feasible Solution, $\hat{P}$, has a maximal number of zeros and results in a reducible chain with three communicating classes, namely $00, \{01, \ 10\}, 11$. By convexity $P_u = u \cdot \overline{P} + (1 - u) \cdot \hat{P}$ is also a solution, for any $0 < u < 1$, so that for $u = 0.5$ and $u = 0.2$ we obtain respectively the following two different 2-grams:

$$
\begin{bmatrix}
P_{0.5}(00, 00) & = & 0.839 \\
P_{0.5}(00, 01) & = & 0.161 \\
P_{0.5}(10, 00) & = & 0.339 \\
P_{0.5}(10, 01) & = & 0.661 \\
P_{0.5}(01, 10) & = & 1.000 \\
P_{0.5}(01, 11) & = & 0.000 \\
P_{0.5}(11, 10) & = & 0.500 \\
P_{0.5}(11, 11) & = & 0.500
\end{bmatrix}
,
\begin{bmatrix}
P_{0.2}(00, 00) & = & 0.936 \\
P_{0.2}(00, 01) & = & 0.064 \\
P_{0.2}(10, 00) & = & 0.136 \\
P_{0.2}(10, 01) & = & 0.864 \\
P_{0.2}(01, 10) & = & 1.000 \\
P_{0.2}(01, 11) & = & 0.000 \\
P_{0.2}(11, 10) & = & 0.200 \\
P_{0.2}(11, 11) & = & 0.800
\end{bmatrix}
.
$$

---

[1] $\hat{P}$ is a Basic Feasible Solution obtained by employing the Matlab linprog function.

4) Now compare the original 2-order statistics specified by $M$ with the statistics specified by the two new models, namely $P_{0.5}$ and $P_{0.2}$ as above:

$$
\begin{bmatrix}
R(00) & = & 0.513 \\
R(01) & = & 0.244 \\
R(10) & = & 0.244 \\
R(11) & = & 0.000
\end{bmatrix}
,
\begin{bmatrix}
R_{0.5}(00) & = & 0.513 \\
R_{0.5}(01) & = & 0.244 \\
R_{0.5}(10) & = & 0.244 \\
R_{0.5}(11) & = & 0.000
\end{bmatrix}
,
\begin{bmatrix}
R_{0.2}(00) & = & 0.513 \\
R_{0.2}(01) & = & 0.244 \\
R_{0.2}(10) & = & 0.244 \\
R_{0.2}(11) & = & 0.000
\end{bmatrix}
.
$$

They are numerically identical as expected. Finally we verify that the 3-order statistics are all different from each other and from the 3-order statistics of the original data, $R$, previously listed.

$$
\begin{bmatrix}
\overline{R}(000) & = & 0.348 \\
\overline{R}(001) & = & 0.165 \\
\overline{R}(010) & = & 0.244 \\
\overline{R}(011) & = & 0.000 \\
\overline{R}(100) & = & 0.165 \\
\overline{R}(101) & = & 0.079 \\
\overline{R}(110) & = & 0.000 \\
\overline{R}(111) & = & 0.000
\end{bmatrix}
,
\begin{bmatrix}
\hat{R}(000) & = & 0.513 \\
\hat{R}(001) & = & 0.000 \\
\hat{R}(010) & = & 0.244 \\
\hat{R}(011) & = & 0.000 \\
\hat{R}(100) & = & 0.000 \\
\hat{R}(101) & = & 0.244 \\
\hat{R}(110) & = & 0.000 \\
\hat{R}(111) & = & 0.000
\end{bmatrix}
,
\begin{bmatrix}
R_{0.5}(000) & = & 0.430 \\
R_{0.5}(001) & = & 0.083 \\
R_{0.5}(010) & = & 0.244 \\
R_{0.5}(011) & = & 0.000 \\
R_{0.5}(100) & = & 0.083 \\
R_{0.5}(101) & = & 0.161 \\
R_{0.5}(110) & = & 0.000 \\
R_{0.5}(111) & = & 0.000
\end{bmatrix}
,
\begin{bmatrix}
R_{0.2}(000) & = & 0.480 \\
R_{0.2}(001) & = & 0.033 \\
R_{0.2}(010) & = & 0.244 \\
R_{0.2}(011) & = & 0.000 \\
R_{0.2}(100) & = & 0.033 \\
R_{0.2}(101) & = & 0.211 \\
R_{0.2}(110) & = & 0.000 \\
R_{0.2}(111) & = & 0.000
\end{bmatrix}
.
$$

These 3-order statistics are calculated using the relationships

$$\tilde{R}(ayb) = R(ay) \cdot \tilde{P}(ay, yb)$$

for the various $\tilde{R}$, $a$, $y$, $b$. Moreover, the $R_u$ are the same convex combinations as the the various $P_u$'s. This example illustrates the various constructions we have described in complete generality in the previous section.

## V. A Covert Channel Coding Technique

In the previous section, we showed that given observed string frequencies, $R(z)$, $z \in \Sigma^k$ we can construct multiple Markov Chains, $M$, whose states are the $k$-grams ($z \in \Sigma^k$), transition probabilities are $P(ay, yb)$, $a \in \Sigma$, $y \in \Sigma^{k-1}$ and whose stationary distributions are precisely the observed $R$.

We now show how to use such a Markov Chain to encode messages while preserving the statistics, $R$, of the channel. This means that someone monitoring the channel will observe the same $k$-gram statistics in spite of the fact that covert messages can be communicated within that channel. As noted before, this can be exploited by either attacker or defender.

Conceptually, the coding concept is the opposite of the classical Shannon Source Coding Theorem [17] in the sense that traditionally we start with a stochastic source with entropy rate $H$ that we seek to compress into binary strings whereas in this case we start with a collection of $2^r$ messages which we wish to efficiently encode using the dynamics and statistics of the given stochastic process. Because we have to respect the statistics of the channel, the encoding will typically not be be compressing but expanding the number of bits needed. Nonetheless, we still seek efficiency with respect to observing the channel's $k$-gram statistics.

In this work, we will assume, for simplicity, that the communication covert channel is *noiseless* noting that the results can be extended to noisy channels in the traditional way. A more thorough analysis is deferred to a future study in which the Shannon capacity of noisy channels will be considered.

This construction involves several steps:

1) Compute the entropy of the irreducible Markov Chain $M$, $H_M$, specified by transition probabilities, $P_M$, and stationary distribution, $R_M$:

$$H_M = - \sum_{ay \in \Sigma^k} R_M(ay) \sum_{b \in \Sigma} P_M(ay, yb) \log_2(P_M(ay, yb)).$$

Note that we construct the Markov Chains to have a given stationary distribution, $R$, so only $P_M$ is different for the different models.

For the examples developed in the previous section, we have computed:

$$H_{\overline{P}} = 0.6863 \;,\; H_{\hat{P}} = 0 \;,\; H_{P_{0.2}} = 0.3165 \;,\; H_{P_{0.5}} = 0.5520.$$

Note that $\hat{P}$ is entirely deterministic and so has zero entropy.

Since we constructed these Markov Chains so that different transitions from a state correspond to different observables (that is, be DPFA's), knowledge of the initial state of the Markov Chain results in a one-to-one correspondence between state sequences and observation sequences. Hence the entropy rates of both the Markov Chain state sequences and resulting observation sequences are the same.

Let $\mathcal{Y}_s$ represent the stochastic process of observations produced by the constructed Markov Chain, $M$, starting in state $s \in M$. All states in $M$ are recurrent by construction so the entropy rate of each process $\mathcal{Y}_s$ is the same and equal to $H_M$.

2) Apply the Shannon-MacMillan-Breiman Asymptotic Equipartition Property (AEP) Theorem [17] to each $\mathcal{Y}_s$ showing that for large $n$ there are approximately $2^{nH_M}$ *typical* sequences of length $n$ of $\mathcal{Y}_s$ and each occurs with probability approximately $(1/2)^{nH_M}$. Consequently, in order to encode $2^r$ covert messages, say $C_i$ with $1 \le i \le 2^r$ we must have $r \le nH_M$ or equivalently $n \ge r/H_M$ so $n$ is selected to encode $2^r$ different covert message sequences accordingly.

3) Construct length $n$ typical sequences of $\mathcal{Y}_s$ by starting in state $s$ and then performing a random walk of length $n$ in $M$ according to the probabilities $P_M$. Such random walks define observation sequences of length $n$ in $\Sigma^n$. Produce $2^r \le 2^{nH_M}$ unique sequences for each state $s$, labeling them as $Y_s(i)$ where $1 \le i \le 2^r \le 2^{nH_M}$. (If a random walk produces a sequence already generated, simply repeat until a novel random walk is produced.)

4) Note that the $k$-gram frequencies of each $z \in \Sigma^k$ within the $Y_s(i)$ approach the original $R(z)$ as $n \to \infty$ because $R$ is the stationary distribution of the Markov Chain and $Y_s(i)$ is produced by taking a random walk in the chain.

5) For each state, $s$, assign the covert message $C_i$ to $Y_s(i)$. Pick a random initial state $s(0)$ and assign a sequences of covert messages $C_{i_1} C_{i_2} ... C_{i_m}$ to

$$Y_{s(0)}(i_1) Y_{s(1)}(i_2) ... Y_{s(m-1)}(i_m)$$

where $s(j)$ is recursively defined as the state in which $Y_{s(j-1)}(i_j)$ ended.

Because each random walk in the sequence thus constructed starts in the state in which the previous random walk ended, the concatenated sequence of random walks is also a legal random walk in the Markov Chain, obeying all the transition probabilities. Moreover, the $k$-gram statistics in the overall concatenated sequence of $mn$ observations is approximately $R$ and approaches $R$

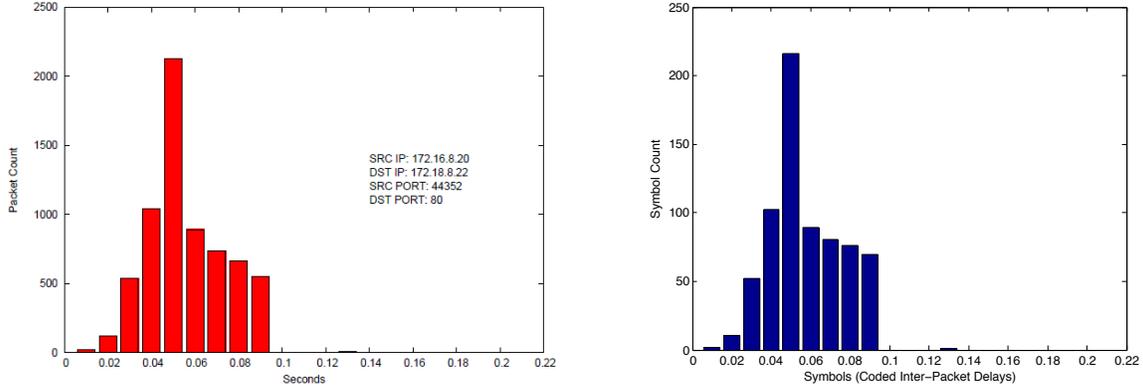

Fig. 4.    The frequencies of the 13 different delays measured from the codeword sequence are in the right graph. This is to be compared with the left graph which is from the empirical data as in Figure 2. Someone monitoring the delays would see no change in the distribution but a covert channel is present.

as $n \to \infty$. The encoded sequence is uniquely decodable by the receiver as well.

To illustrate this construction, consider the example presented in the left of Figure 2 where we use the 1-order statistics as in equation (2). We take the convex combination (see Section II)

$$P = 0.75 \cdot P_B + 0.25 \cdot P_D,$$

which results in an entropy of $H_P = 0.004$ as computed from (9). We build a $(2^r, n)$ codebook as described above with $r = 8$ and $n = \lceil r/H_P \rceil = 1995$. That is, this encodes binary sequences of length 8 into inter-packet delays of length 1995. We encoded 16 blocks of 8 random source bits each into $16 \cdot 1995 = 31920$ symbols from the alphabet $\Sigma = \{1, 2, \ldots, 13\}$ which correspond to the delays in the left graph of Figure 2.

The obtained 1-order statistics of the resulting 31920 long concatenated codeword are

$$R' = [0.0029\ 0.0144\ 0.0734\ 0.1453\ 0.3094\ 0.1295\ 0.1151\ 0.1079\ 0.1007\ 0\ 0\ 0\ 0.0014];$$

and are depicted in Figure 4. Note the empirical frequencies and graphs are identical to the displayed precision.

This illustrates empirically the effectiveness of the construction described in this paper. Matlab code for reproducing these results is available upon request.

## VI. CONCLUSIONS AND FUTURE WORK

This paper has demonstrated that covert channels can exist even when arbitrarily high order statistics about a channel are estimated and monitored. The resulting covert channels can be used to either exploit or defend the channel and the advantage goes to the party that has the ability to estimate the highest order statistics.

The adversarial nature of this situation falls within the scope of *cognitive attacks* [26], [27]. It can be described in abstract as follows: the environment (for example, inter-packet delays) is modeled as a stochastic process $\mathcal{X}$ (such as a Hidden Markov Model, Markov Chain or other formalism). Both the attacker, $A$, and the defender, $D$, monitor the environment through functions $f_A \in \mathcal{F}$ and $f_D \in \mathcal{F}$ respectively (for example, $f_D(\mathcal{X})$ could be the probability distribution of $k$-grams produced by $\mathcal{X}$).

The attacker guesses $f_D$ and manipulates $\mathcal{X}$ in order to produce a new process, $A(\mathcal{X})$, so that covert communications can be performed while respecting the behavior that the defender expects; namely, $f_D(\mathcal{X}) = f_D(A(\mathcal{X}))$.

On the other hand, the defender, by anticipating the attacker's guess of $f_D$, picks a different $\tilde{f}_D$ and manipulates $\mathcal{X}$ to produce a new process $D(\mathcal{X})$ so that:

1) $f_D(A(D(\mathcal{X}))) = f_D(D(\mathcal{X})) = f_D(\mathcal{X}) = f_D(A(\mathcal{X}))$: the defensive shaping action is imperceptible to the defender who uses $f_D$;

2) $\tilde{f}_D(A(D(\mathcal{X}))) \neq \tilde{f}_D(D(\mathcal{X}))$: the attacker's action (that is, creation of a covert channel) is detectable by the defender.

The game consists of attacker and defender guessing and then exploiting each other's monitoring strategy and manipulating the environment accordingly. The common objective of the players is to alter the environment in a manner that would be imperceptible to the opponent in order to perform a secret task (covert communication or covert channel detection).

This work raises some questions which are deferred to future work. In particular, the following directions are worthy of future investigation:

- Inter-packet delays involve real-world time so the question of stability when shaping the channel must be considered. That is, packets can be delayed by certain times only if there are packets in the queue to be delayed. Discussions of such queuing aspects of timing channels and the possibility of jamming them have been studied [28]. Relating this work to timing channel jamming will be investigated.

- We used a circular buffer in Section IV to numerically estimate $k$-gram statistics so that the statistics have the required marginalization properties. A single pass, online algorithm for implementing this circular buffer only requires storing the first and last $k$ symbols of the data. In the absence of such a buffer, the empirical statistics will not in general obey the marginalization identities and so some additional processing would be required. The use of singular value decompositions, non-negative matrix factorizations or other decomposition methods for imposing the regularity might be worth exploring further as alternatives to the circular buffer approach.

- In principle, one can attempt to build automata smaller than the Markov Chains we construct. In particular, Probabilistic Finite Automata (PFA) [29], [30], [31] could implement Markov Chains based on $k$-grams but using fewer states. Unlike $k$-gram based Markov Chains, $k$-PSAs have states that are labeled with input sequences of length at most $k$. So they can be seen as "variable length" $k$-gram Markov Chains. They can be learned efficiently in the KL-PAC sense [32], [33], [34] and are generally smaller than $k$-gram based Markov chains (by having fewer states).

- Within the space of possible Markov Chains that realize given $k$-gram statistics, it would be good to select the "best" chain from the point of maximizing entropy so that the covert channel coding is as efficient as possible. Our experiments suggest that the so-called Standard Solutions presented in Section III-A have the largest entropy although we have not been able to prove that analytically.

- It is reasonable to ask how our results relate to the use of Hidden Markov Models for modeling traffic, as for example in [3]. It is known that a Hidden Markov Model with $n$ states is completely determined by the $2n$-grams produced by the model so that reproducing $2n$-gram statistics will result in the same $n$ state Hidden Markov Model [8].

**Valentino Crespi** received his Laurea Degree and his Ph.D. Degree in Computer Science from the University of Milan, Italy, in July 1992 and July 1997, respectively. From September 1998 to August 2000 he was an Assistant Professor of Computer Science at the Eastern Mediterranean University, Famagusta, North Cyprus and from September 2000 to August 2003 he worked at Dartmouth College, Hanover, NH, as a Research Faculty. Since September 2003 he has been on faculty at the Department of Computer Science, California State University, Los Angeles, currently in the capacity of Associate Professor. His research interests include Distributed Computing, Tracking Systems, UAV Surveillance, Sensor Networks, Information and Communication Theory, Complexity Theory and Combinatorial Optimization. At Dartmouth College he developed the TASK project and consulted to the Process Query Systems project, directed by Prof. George Cybenko. At CSULA he has been teaching lower division, upper division and master courses on Algorithms, Data Structures, Java Programming, Compilers, Theory of Computation and Computational Learning of Languages and Stochastic Processes. During his professional activity Dr Crespi has published a number of papers in prestigious journals and conferences of Applied Mathematics, Computer Science and Engineering. Moreover Dr Crespi is currently a member of the ACM and of the IEEE.

**George Cybenko** is the Dorothy and Walter Gramm Professor of Engineering at the Thayer School of Engineering at Dartmouth College. Prior to joining Dartmouth, he was Professor of Electrical and Computer Engineering at the University of Illinois at Urbana-Champaign. His current research interests are in machine learning, signal processing and computer security. Cybenko was founding Editor-in-Chief of IEEE Computing in Science and Engineering and IEEE Security & Privacy. He earned his B.Sc. (Toronto) and Ph.D (Princeton) degrees in mathematics and is a Fellow of the IEEE.

**Annarita Giani** received her Laurea (Master degree) in Mathematics from the Università di Pisa, Italy. Thereafter, she worked as a researcher for the Italian Registration Authority, as well as the Istituto di Informatica e Telematica del Consiglio Nazionale delle Ricerche in Pisa. In 2001 she moved to the United States to commence a Ph.D in Computer Engineering at Dartmouth College's Thayer School of Engineering, Hanover, New Hampshire. While at Dartmouth, she participated to the Process Query System (PQS) project sponsored by the Advanced Research and Development Activity (ARDA). Her dissertation addressed issues relating to computer security, anomaly tracking and cognitive attacks. She received her doctoral degree in 2007. She presently holds the position of postdoctoral fellow at the Department of Electrical Engineering and Computer Science at the University of California at Berkeley. While at Berkeley, she has been working on security for wireless and sensor networks, as well as security issues related to body sensor networks and critical infrastructures.